%% file: root.tex
\documentclass[journal]{IEEEtran}

\usepackage[T1]{fontenc}
\usepackage[utf8]{inputenc}
\usepackage{textcomp}
\usepackage{xcolor}
\usepackage{algorithm}
\usepackage{algorithmic}
\usepackage{amsmath}
\usepackage{cite}
\usepackage{amssymb}
\usepackage{graphicx}
\usepackage{overpic}
\usepackage{stfloats}
\usepackage{etoolbox}
\usepackage{balance}

\renewcommand{\IEEEPARstartWORDCAPSTYLE}{\relax}

\makeatletter
\def\@IEEEBIOskipN{0.2\baselineskip}
\def\@IEEEBIOphotodepth{1.0in}
\def\@IEEEBIOhangdepth{1.0in}
\def\@IEEEBIOhangwidth{0.88in}
\makeatother

\title{A 2D Image Feedback-Based Autonomous Endovascular Navigation Method: A Perception and Control Framework}

\author{Sheng Cao, Shuxiang Guo, \IEEEmembership{Fellow, IEEE}, Yonggan Yan,\\
Jian Guo, Chunying Li, Bin Wang, Mingchao Ding,\\
Yongxin Zhang, Yongwei Zhang, Pengfei Yang, and Jianmin Liu%
\thanks{Research supported by the Shenzhen ``Medical Engineering Integration'' Special Support Program Project under Grant F-2024-Z99-502365, Shenzhen, China; in part by the High level of special funds from Southern University of Science and Technology under Grant G03034K003, Shenzhen, China. (Corresponding author: Shuxiang Guo.)}%
\thanks{S. Cao is with the Beijing Institute of Technology, Beijing 100081, China, and also with the Advanced Institute for Ocean Research, Southern University of Science and Technology, Shenzhen, 518055, China, and also with the Shenzhen Institute of Advanced Biomedical Robot Company Limited, Shenzhen 518000, Guangdong Province, China.}%
\thanks{S. Guo is with the Department of Electronic and Electrical Engineering, Southern University of Science and Technology, Shenzhen, Guangdong 518055, China, and also with the School of Life Science and the Key Laboratory of Convergence Medical Engineering System and Healthcare Technology, Ministry of Industry and Information Technology, Beijing Institute of Technology, Beijing 100081, China (e-mail: guo.shuxiang@sustech.edu.cn).}%
\thanks{Y. Yan is with the School of Mechanical, Electronic and Control Engineering, Beijing Jiaotong University, Beijing, China.}%
\thanks{J. Guo is with the Shenzhen Institute of Advanced Biomedical Robot Company Limited, Shenzhen 518000, Guangdong Province, China.}%
\thanks{C. Li is with the Southern University of Science and Technology, Shenzhen, Guangdong 518055, China.}%
\thanks{B. Wang and M. Ding are with Aerospace Center Hospital and the Key Laboratory of Convergence Medical Engineering System and Healthcare Technology, Beijing Institute of Technology, Beijing 100081, China.}%
\thanks{J. Liu, P. Yang, Y.-X. Zhang, and Y.-W. Zhang are with the Neurovascular Center, Changhai Hospital, Naval Medical University, Shanghai, China.}%
}

\begin{document}

\maketitle

\color{black}

\begin{abstract}
In endovascular interventions, physicians rely on planar DSA and X-ray fluoroscopy to interpret vessel anatomy and device motion, yet converting these images into feedback for autonomous navigation remains difficult. A perception and control method is formulated for two-dimensional image feedback. Vessel masks are converted into a labeled centerline graph, and device-tip estimates are projected onto a planned path to obtain progress, deviation, and direction. A rule-based controller uses these quantities for delivery, branch selection, probing, rollback, and recovery. The method was evaluated on a physical dual-device robotic platform in 65 closed-loop runs using simulator-rendered planar images across three paths. All runs reached at least 0.95 maximum observed guidewire progress; two runs (3.1\%) met the prespecified failure rule of three or more autonomous stop events. The results establish route-level performance on the tested platform; clinical image sequences and broader anatomy remain to be evaluated.

\end{abstract}

\section{Introduction}
\input{intro}

\section{System Overview and Sensing Design}
\label{sec:system_overview}
\begin{figure}[!t]
\centering
\includegraphics[width=\linewidth]{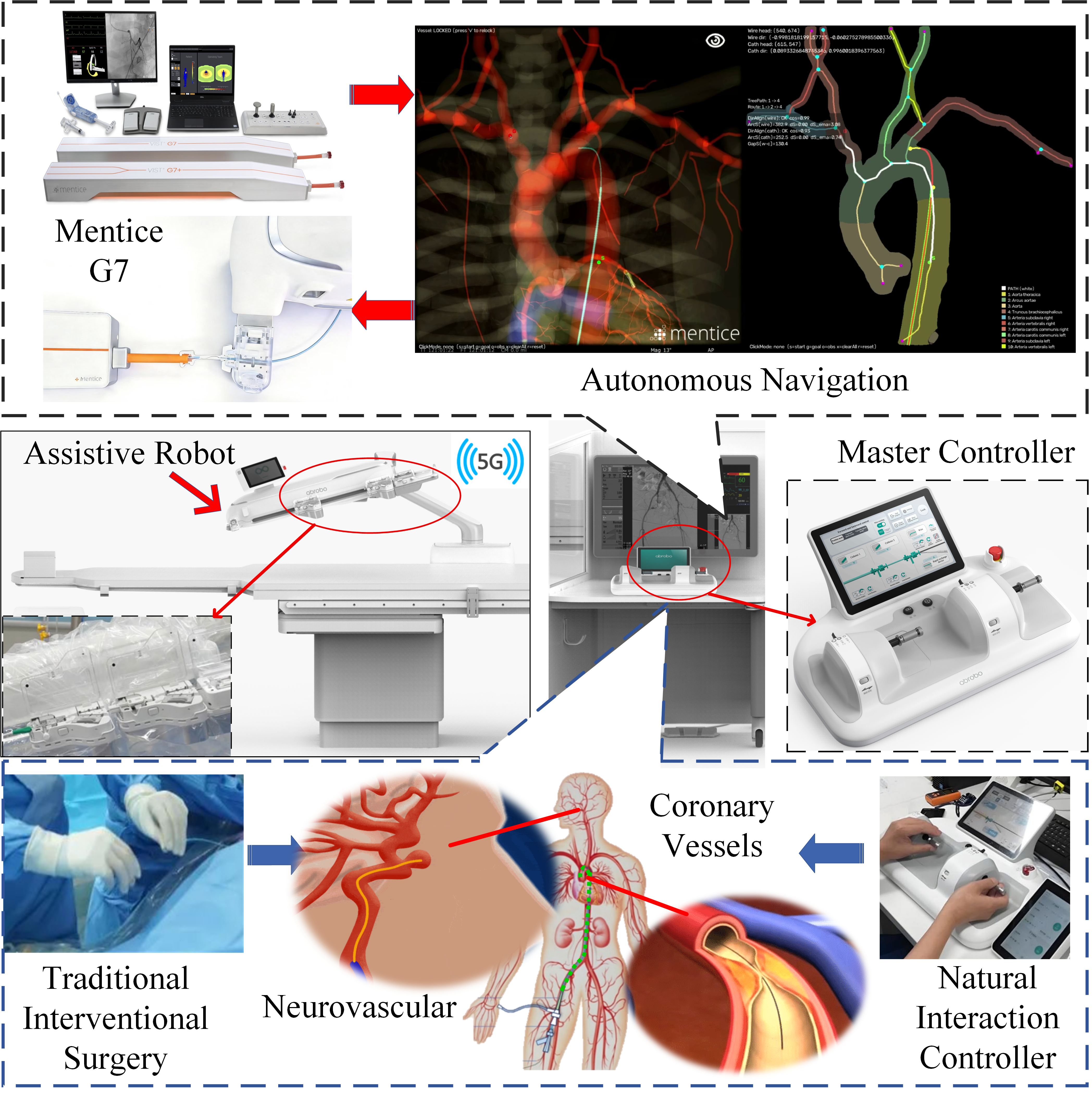}
\caption{Overall system and human-in-the-loop workflow. Mentice provides the planar image stream, the physical dual-device robot executes autonomous commands, and the physician supervises and can override the process.}
\label{fig:overall_system}
\end{figure}

The platform combines the Mentice VIST G7 vascular simulator, a physical dual-device robot, the autonomous-navigation workstation, and a physician-side controller (Fig.~\ref{fig:overall_system}). Mentice supplies the planar projection; the robot changes the guidewire and catheter configuration; and the workstation estimates the next visual state before issuing a command. The physician can supervise or take over the robot. This sequence closes the image--state--action loop.

The workstation converts each registered Mentice frame into a state relative to the planned path, and the controller uses that state to select a primitive for the physical robot. The measurements and outcomes reported below therefore describe a supervised robotic experiment with simulator-provided anatomy; no clinical image acquisition is included.

Mentice provides a fixed planar projection. The vessel representation is initialized once and locked, while the physical robot changes the device configuration and later frames update the guidewire and catheter masks. The experiment therefore isolates dynamic image measurements from the static route map. It does not require three-dimensional reconstruction or force sensing, unlike recent CT-to-2D X-ray registration approaches \cite{saderi_registration_2025,ye_registration_2025}.

Translation and rotation do not determine the final device motion by themselves. Contact, friction, and lumen geometry redirect the flexible guidewire. The controller does not measure these forces directly; it uses successive images to estimate tip displacement, path progress, path distance, direction consistency, and wire--catheter separation. These quantities are the feedback variables for the physical robot.

A substantial portion of recent autonomous endovascular-navigation research has focused on reinforcement learning and other learned policies \cite{rl_zeroshot_guidewire_2024,rl_pathplanning_guidewire_2024,safe_dual_device_rl_2025,autonomous_endovascular_review_2025,embodied_endovascular_review_2026}. The present study addresses route-level navigation with a physical robot and simulator-rendered planar images. Progress, path distance, tangent alignment, and junction proximity define the control state. A deterministic state machine uses these variables for advancement, branch probing, rollback, recovery, and catheter following, while the physician selects the start, target, and risk points. Learned control for local guidewire--catheter coordination remains a separate future problem.

\section{Path-Based Visual Measurement Model}
\label{sec:problem_formulation}
Let $I_t$ denote a planar image acquired from DSA or X-ray fluoroscopy, or generated by the simulator, at time $t$. The task is specified by an operator-selected start point and target point. A vessel centerline graph is extracted from a locked vessel segmentation, and a topology-constrained route $\mathcal{P}=\{\mathbf{r}(s):0\leq s\leq L\}$ connects the two points. The variable $s$ denotes pixel arc length measured from the route start. The projected start and target coordinates are $s_0$ and $s_g$, respectively. The constructed route normally gives $s_0=0$ and $s_g=L$. The sensing problem is to estimate device tips, local directions, and residuals relative to the planned path from $I_t$; the resulting measurements are then consumed by the downstream controller. Only image-derived geometric feedback is used.

Let $\mathbf{p}^{w}_t$ and $\mathbf{p}^{c}_t$ denote the guidewire and catheter tip positions. The corresponding unit image-plane direction vectors are $\mathbf{u}^{w}_t$ and $\mathbf{u}^{c}_t$. Component notation follows $\mathbf{p}^{i}_t=(x^i_t,y^i_t)$ and $\hat{\mathbf{p}}^{i}_t=(\hat{x}^i_t,\hat{y}^i_t)$. Projection onto $\mathcal{P}$ gives
\begin{equation}
s^i_t=\underset{s\in[0,L]}{\arg\min}\,\|\mathbf{p}^i_t-\mathbf{r}(s)\|_2,
\quad \hat{\mathbf{p}}^i_t=\mathbf{r}(s^i_t),\quad
d^i_t=\|\mathbf{p}^i_t-\hat{\mathbf{p}}^i_t\|_2,
\label{eq:path_projection}
\end{equation}
for $i\in\{w,c\}$. The raw increments, target-directed one-step progress, interval progress, and wire--catheter separation are
\begin{equation}
\begin{aligned}
\Delta s^i_t&=s^i_t-s^i_{t-1},\quad i\in\{w,c\},\\
q^w_t&=\operatorname{sgn}(s_g-s^w_{t-1})\Delta s^w_t,\\
Q^w_{t_0,t}&=\operatorname{sgn}(s_g-s^w_{t_0})(s^w_t-s^w_{t_0}),\\
g^{wc}_t&=s^w_t-s^c_t.
\end{aligned}
\label{eq:progress_gap}
\end{equation}
The action set contains guidewire insertion, retraction, and left or right rotation. Catheter insertion and retraction are also available. The next image determines the realized outcome of each command. The controller maps the latest valid geometric observation and internal mode to one action or bounded action sequence. An active off-path recovery first runs to its clearance condition. Otherwise, interaction-risk retreat preempts junction probing and nominal guidewire advancement. Post-target catheter following begins only after guidewire target arrival. This ordering describes the experimental controller and is not a formal safety guarantee.

\section{Image-Based Perception and State Estimation}
\label{sec:visual_state}

\begin{figure*}[t]
\centering
\includegraphics[width=\textwidth]{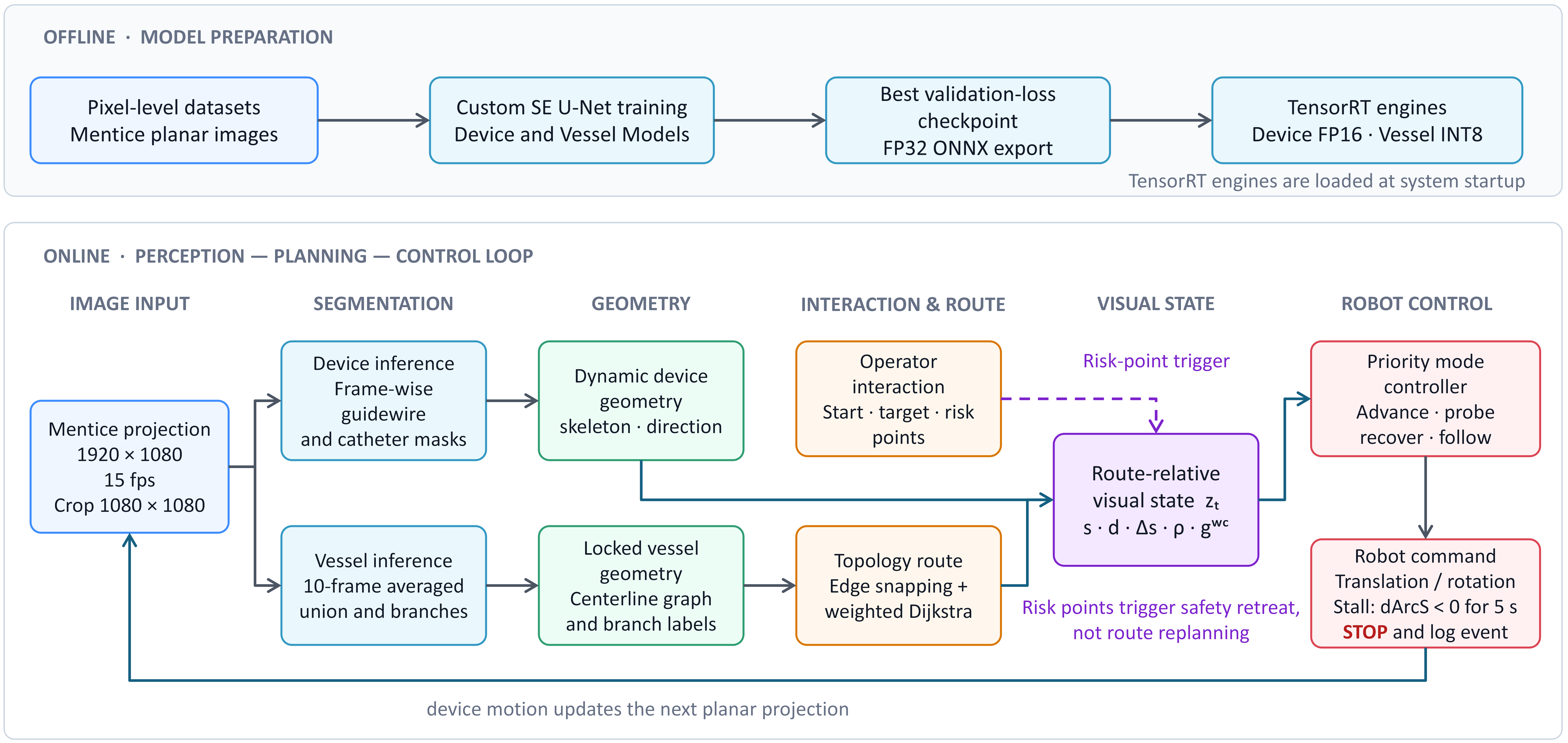}
\caption{System architecture and data flow. Offline, custom device and vessel SE U-Net models are trained, exported to ONNX, and converted to TensorRT engines. Online, device masks provide frame-wise tip and direction estimates, whereas the averaged vessel output is locked as a labeled centerline graph. Operator-selected start and target points define the topology-constrained route. Operator-marked interaction-risk points form a separate early-trigger channel: proximity invokes priority retreat before the delayed off-path condition, without modifying the graph or replanning the route. This heuristic response is not a formal safety guarantee. Projection of both tips onto the route produces the visual state used by the priority controller. A discrete command drives the robot and changes the physical device configuration. Mentice updates the device--vessel interaction and renders the realized motion in the next simulator-rendered image, which closes the visual feedback loop.}
\label{fig:system_dataflow}
\end{figure*}

The complete system architecture and data flow are summarized in Fig.~\ref{fig:system_dataflow}.

Figure~\ref{fig:system_dataflow} separates offline model preparation from online control. The vessel representation is initialized once and locked, whereas guidewire and catheter masks are refreshed frame by frame. Route projection, the risk-point trigger, the controller mode, the robot command, and the subsequent Mentice image form the causal loop used for every logged run. This separation is also why IoU is interpreted as temporal mask stability and not as a substitute for a ground-truth clinical segmentation study.

\subsection{Segmentation-model development and measurement targets}
Two independently trained segmentation models use a custom U-Net with squeeze-and-excitation (SE) blocks. Both models receive RGB Mentice-rendered planar images. The device model targets the two dynamic structures that determine the visual state: guidewire and catheter. Its dataset contains 102 training, 22 validation, and 22 test images, and its two output channels are allowed to overlap where the devices intersect. The vessel model targets the static route reference. Its dataset contains 117 training, 15 validation, and 14 test images; masks were annotated and exported with 3D Slicer. It outputs ten anatomical branch channels and one vessel-union channel, allowing the same representation to support both branch labeling and centerline extraction.

The encoder contains four resolution levels with 64, 128, 256, and 512 channels. Each level uses two $3\times3$ convolutions, BatchNorm2d, ReLU, and an SE block with reduction ratio 8. Max pooling performs downsampling. A 1024-channel bottleneck uses dropout 0.1. The decoder applies bilinear upsampling, $1\times1$ channel reduction, skip concatenation, and the same double-convolution SE block. A final $1\times1$ convolution produces two device logits or eleven vessel logits.

Training uses foreground-centered random $512\times512$ crops and batch size 1. Input intensity is scaled to $[0,1]$. The loss is $0.7\mathcal{L}_{\mathrm{BCE}}+0.3\mathcal{L}_{\mathrm{Dice}}$. AdamW uses an initial learning rate of $10^{-4}$ and weight decay $2\times10^{-4}$. A ReduceLROnPlateau scheduler halves the learning rate after five validation epochs without improvement. Mixed-precision training and gradient clipping at 1.0 are enabled. Both models are trained for 60 epochs with seed 2025. The checkpoint with the lowest validation loss is retained. Training and deployment measurements appear in Sec.~\ref{sec:seg_benchmark}.

\subsection{Online mask generation and temporal locking}
The online perception process receives a $1920\times1080$ YUYV stream at a requested rate of 15 frames/s and uses the right $1080\times1080$ crop. The selected PyTorch checkpoints are exported to FP32 ONNX. TensorRT 10.3 then builds FP16 and INT8 engines with batch size 1, dynamic spatial dimensions from 256 to 2048 pixels, and an optimization size of $640\times640$. Full-frame inference uses $640\times640$ tiles with 64-pixel overlap and Hann-weighted logit fusion. The output logits are converted to binary masks with a threshold of 0.5. The closed-loop experiments use the device FP16 engine and the vessel INT8 engine. Table~\ref{tab:seg_engine_benchmark} reports the accuracy agreement and latency that support this mixed-precision deployment.

The device masks are updated with each processed frame because their temporal stability directly determines tip and direction measurements. Eight-connected components smaller than 200 and 500 pixels are removed from the guidewire and catheter masks, respectively. Components smaller than 2000 pixels are removed from the vessel union mask. To suppress transient exposure and image noise, the vessel model is applied to the average of the first ten frames used for vessel-map initialization. The resulting vessel mask, branch probabilities, skeleton, and graph are then locked for the current view, while device masks and tips remain dynamic. For each robot trial, initialization or re-locking is performed before the locked view is used. The operator can explicitly re-lock the vessel representation if the simulated view changes.

\subsection{Device-tip measurement and local-direction estimation}
For each device mask, the largest connected component is retained and skeletonized. A two-sweep geodesic search is performed on a 0.4-scale mask. A distance-transform maximum supplies an interior seed. The farthest foreground point from the seed defines the first endpoint. The farthest point from the first endpoint defines the second. Both endpoints are mapped to the original resolution. The upper endpoint is selected as the tip under the fixed experimental image orientation.

Starting at the tip, up to 30 skeleton points are traced toward the device body. Principal-component fitting estimates the local image-plane direction, with its sign oriented from the traced tail toward the tip. A directional exponential filter with update coefficient 0.25 reduces frame-to-frame fluctuations. The per-frame tip displacement and its root-mean-square value over a 30-frame window are logged as detection-stability measures, while interframe mask IoU provides a complementary temporal-stability measure. The upper-endpoint rule and the locked vessel representation are view-dependent assumptions of the reported experiments.

\subsection{Topology-constrained route construction}
The locked vessel mask is skeletonized into a one-pixel centerline. Under eight-connectivity, skeleton pixels with degree one are endpoints and pixels with degree at least three are junctions. These pixels form the node set of $\mathcal{G}=(\mathcal{V},\mathcal{E})$; tracing the skeleton between successive nodes forms graph edges. Each edge stores its ordered centerline pixels and accumulated Euclidean length, with unit axial steps and $\sqrt{2}$ diagonal steps. Vessel-network branch probabilities are sampled along each edge, and the largest mean probability assigns its anatomical branch label.

The operator marks the route start and target in the current planar image. Both clicks are projected onto the nearest graph-edge polylines. Temporary nodes split the corresponding edges. A label tree supplies the anatomically allowed branch labels between start and target. Weighted Dijkstra search then uses
\begin{equation}
C(e)=
\begin{cases}
\ell(e), & \lambda(e)\in\Lambda_{sg},\\
8\ell(e), & \lambda(e)\notin\Lambda_{sg},
\end{cases}
\label{eq:topology_edge_cost}
\end{equation}
where $\ell(e)$ is the edge length and $\lambda(e)$ is the branch label. The set $\Lambda_{sg}$ contains the allowed labels between the selected branches. The soft eightfold penalty preserves reachability when a local label is missing. It also favors the anatomically consistent route.

The ordered graph edges are concatenated into the polyline $\mathcal{P}$. Junction nodes within 10 pixels of this polyline are projected onto it and sorted by arc length. Both device tips are then projected segment by segment according to (\ref{eq:path_projection}). The vision process also computes
\begin{equation}
\bar{\Delta s}^{i}_t=0.75\bar{\Delta s}^{i}_{t-1}+0.25\Delta s^i_t,
\quad
\rho^i_t=\langle\mathbf{u}^i_t,\boldsymbol{\tau}(s^i_t)\rangle,
\label{eq:ema_alignment}
\end{equation}
where $i\in\{w,c\}$. The unit vector $\boldsymbol{\tau}$ is a local route tangent estimated over neighboring polyline samples. Thus, $\rho^i_t\in[-1,1]$ is a direction cosine. The vision interface displays a Boolean indicator based on $\rho^i_t\geq0.5$. The controller uses the continuous value $\rho^w_t$ only for junction probing and stagnation checks.

In addition to the start and target, the operator may mark image-plane interaction-risk points on an undesired branch or region. Continued advancement near a marked point is undesirable. The points do not remove graph edges or modify the Dijkstra cost. A separate published set triggers the recovery behavior in Sec.~\ref{sec:interaction_recovery} when the guidewire approaches a point.

The published visual state is
\begin{equation}
\label{eq:strategy_state_vector}
\begin{aligned}
\mathbf{z}_t=\big[&\mathbf{p}^w_t,\mathbf{p}^c_t,\mathbf{u}^w_t,\mathbf{u}^c_t,
s^w_t,s^c_t,\Delta s^w_t,\\
&\bar{\Delta s}^{w}_t,d^w_t,d^c_t,\rho^w_t,\rho^c_t,g^{wc}_t\big].
\end{aligned}
\end{equation}
The state message also contains the start and goal coordinates and arc-length projections. Junction coordinates, junction arc lengths, risk points, validity flags, and raw squared projection distances are included. The control process derives $q_t^w$, $Q^w_{t_0,t}$, the nearest junction, the nearest risk point, and Euclidean start and goal distances. The interface therefore converts the image stream into an inspectable observation relative to the planned path before autonomous action selection.

\subsection{Robot command interface}
The control process sends protocol write frames over two CAN interfaces through an SSH-connected gateway. Guidewire and catheter translation use the right and left delivery axes, respectively. One translation primitive is a blocking grip--move--stop--release sequence with a nominal displacement of 14,000~$\mu$m, speed parameter \texttt{0x30}, and 0.30-s inter-step interval. A guidewire rotation primitive uses a protocol step of 100,000 at speed parameter 100. Translation set-point frames are transmitted twice with a 0.02-s interval to reduce packet-loss sensitivity; rotation frames are transmitted once. These values are command-sequencing parameters rather than measured communication or end-to-end response delays. A new decision uses the next valid visual state after the preceding blocking primitive returns.

\section{Path-Guided Control Implementation}
\label{sec:controller}
The path-guided control implementation is a deterministic hierarchical finite-state machine (HFSM) driven by the measured state in (\ref{eq:strategy_state_vector}). It uses mutually exclusive modes and explicit transitions; no learned policy is used. The modes respond to image-observed outcomes of device--vessel interaction. Guidewire navigation contains \emph{Risk-Point Recovery}, \emph{Off-Path Recovery}, \emph{Junction Probing}, and \emph{Auto-Advance}. The post-target \emph{Catheter-Follow} mode begins after guidewire target arrival. A separate operator-triggered reset returns both devices toward the selected start configuration. Reset is not an autonomous navigation mode.

\subsection{Auto-advance, goal detection, and junction stopping}
In \emph{Auto-Advance}, a guidewire translation command becomes eligible no more frequently than every 0.35~s. This is a minimum scheduling gate; the realized command interval also includes completion of the blocking motion primitive and acquisition of a new valid visual state. The sign of $s_g-s^w_t$ selects insertion or retraction toward the target. The reported routes are ordered from start to target. The guidewire reaches the target when $|s^w_t-s_g|\leq1$ pixel. A 25-pixel Euclidean tip-to-target threshold is used if route projection is unavailable.

Before each nominal step, the controller checks the projected junction coordinates. The symmetric junction half-window is
\begin{equation}
h_t=\min\!\left(60,15+10|\bar{\Delta s}^{w}_t|\right)
\quad\text{pixels}.
\label{eq:junction_window}
\end{equation}
The adaptive window tolerates discrete projection jumps. Reaching a junction stops nominal advancement and starts \emph{Junction Probing}. After a successful probe, the same junction is ignored for 2~s and until the tip has progressed 12 arc-length pixels beyond it.

Target-directed progress $q_t^w$ is monitored over 5~s. A tortuous vessel segment may produce little image-plane arc-length change. Nonpositive progress alone therefore does not stop advancement. Motion continues when $d^w_t\leq35$ pixels and $\rho^w_t\geq-0.30$. Motion also continues when direction is unavailable. Otherwise, advancement stops. Probing is requested if the tip is also within 25 pixels of a junction. Persistent $q_t^w<-2$ pixels indicates motion away from the target and stops auto-advance.

\subsection{Deviation correction and off-path recovery}
Local turn correction is driven by projection distance, not by the direction cosine. It enters when $d^w_t\geq22$ pixels and releases when $d^w_t\leq16$ pixels. In the fixed experimental image orientation, the sign of $\hat{x}^w_t-x^w_t$ selects right or left rotation. Rotations are separated by at least 1.10~s; after three consecutive corrections, rotation is inhibited for 3~s to suppress oscillation. This image-axis rule is specific to the fixed view used in the experiments.

A larger deviation invokes \emph{Off-Path Recovery}. If $d^w_t\geq60$ pixels persists for 1.5~s, nominal advancement is paused. Retraction primitives become eligible at a minimum 0.35-s scheduling interval and continue until $d^w_t\leq35$ pixels. The controller then returns to the previously enabled auto-advance logic.

\subsection{Alternating junction scan, insertion test, and rollback}
\label{sec:fork_probing}
At a route junction, rotation alone cannot determine which branch the flexible guidewire will enter because the realized direction depends on subsequent wall contact, friction, and lumen constraint. The controller therefore scans discrete rotation targets relative to the initial orientation in the order $+1,-1,+2,-2,\ldots,+6,-6$, where one unit is one 100,000-unit rotation primitive. At each completed target, it evaluates the score
\begin{equation}
J=d^w_t+22(1-\rho^w_t)+0.25|k|,
\label{eq:probe_score}
\end{equation}
where $k$ is the discrete rotation target. The coefficients form an empirical weighted score; the three terms represent path distance, angular inconsistency, and rotation effort. A five-step insertion test starts when $d^w_t\leq28$ pixels and $\rho^w_t\geq0.05$. The reported routes are oriented from start to target, so positive $\rho^w_t$ points along the planned route. A test also starts after two evaluated targets. This fallback prevents indefinite scanning under noisy orientation estimates.

The sign of $s_g-s^w_t$ determines the insertion direction. Let $t_0$ denote the start of the five-step test. Limited advancement permits the selected rotation and device--vessel interaction to generate a candidate entry. Subsequent images reveal the outcome relative to the planned path. A value of $d^w_t\geq100$ pixels causes immediate rollback. After five steps, strict acceptance requires $Q^w_{t_0,t}\geq0.6$ arc-length pixels, $d^w_t\leq18$ pixels, and $\rho^w_t\geq0.05$. If strict acceptance fails, the tolerance rule accepts $d^w_t\leq45$ pixels. This rule addresses unreliable arc-length and direction estimates in tortuous projections. Otherwise, one inverse translation primitive is executed for each completed test step. The alternating scan then resumes. Matched rollback reverses the commanded translation count but does not guarantee restoration of the same mechanical configuration. A probe episode ends after 100~s or after six consecutive rotation evaluations with $d^w_t\geq100$ pixels.

\subsection{Operator-marked interaction-risk recovery}
\label{sec:interaction_recovery}
The operator may mark interaction-risk points separately from the start and target. A point can designate an undesired branch or region. Continued advancement in that region is interrupted early. During auto-advance or probing, a tip distance below 22 pixels activates recovery. Retraction primitives become eligible at a minimum 0.10-s scheduling interval and continue until the distance reaches 35 pixels. A 25,000-unit rotation follows each of the first two retractions. The horizontal image displacement determines the rotation direction away from the point. The interrupted auto-advance or probing mode then resumes. An already active off-path recovery continues until its own clearance condition. The point-proximity trigger otherwise acts before a new off-path fallback, which requires $d^w_t\geq60$ pixels for 1.5~s. Risk points limit continued entry into a known undesired region. Normal device--vessel contact remains inherent to navigation. The points are not detected obstacles, wall-contact measurements, or a formal safety layer.

\subsection{Post-target catheter following and manual reset}
The controller uses a staged wire-first strategy. Catheter advancement begins only after guidewire target arrival. A catheter command becomes eligible at a minimum 0.40-s scheduling interval after the preceding primitive returns. The catheter-follow controller declares its terminal relation when
\begin{equation}
8.5\leq g^{wc}_t\leq11.5\quad\text{pixels},
\label{eq:catheter_terminal_gap}
\end{equation}
corresponding to a target separation of $10\pm1.5$ arc-length pixels. This interval defines the controller's completion condition for the catheter-follow stage. The stage also terminates if the separation falls below the interval, the route state is lost, 20~s elapse, or 60 catheter steps are reached. Catheter retraction is not used to regulate an undersized gap.

For repeated experiments, an operator-triggered reset first retracts the catheter toward the start. Guidewire reset then uses three phases. Coarse retraction moves the tip to within 60 pixels of the start. PID-regulated translation reduces the vertical error to 10 pixels. A 16-step rotation scan selects the orientation with the smallest horizontal start-point error. This routine standardizes the initial configuration but is not invoked autonomously during target navigation.

\subsection{Online decision sequence}
Algorithm~\ref{alg:arc_length_nav} summarizes the implemented priority and the bounded junction trial.
\begin{algorithm}[t]
\caption{Implemented path-guided navigation controller}
\label{alg:arc_length_nav}
\begin{algorithmic}[1]
\REQUIRE Latest visual state $\mathbf{z}_t$ and internal mode
\IF{operator reset requested}
  \STATE retract catheter; execute three-phase guidewire reset
\ELSIF{off-path recovery active}
  \STATE retract until $d^w_t\leq35$ px
\ELSIF{risk-point recovery active or triggered}
  \STATE retract and apply at most two small rotations until clearance
\ELSIF{off-path recovery triggered}
  \STATE activate recovery; retract until $d^w_t\leq35$ px
\ELSIF{junction probing active or requested}
  \STATE scan rotation targets; execute five-step test
  \STATE accept by implemented strict/tolerance rule, or roll back executed steps
\ELSIF{auto-advance enabled}
  \STATE check goal, junction, deviation, and persistent progress conditions
  \STATE optionally rotate; issue one target-directed translation step
\ELSIF{catheter follow active}
  \STATE advance catheter until (\ref{eq:catheter_terminal_gap}) or a stop condition
\ENDIF
\end{algorithmic}
\end{algorithm}

\section{Experiments}
\label{sec:experiments}

\subsection{Perception training and deployment benchmark}
\label{sec:seg_benchmark}
Table~\ref{tab:seg_training} summarizes the custom SE U-Net training records. Validation Dice is computed against the manual masks. Device values are reported as guidewire/catheter. Vessel values are reported as all channels/anatomical branches/vessel union.

\begin{table}[H]
\caption{Segmentation training summary}
\label{tab:seg_training}
\centering
\begin{tabular}{lcc}
\hline
Metric & Device & Vessel\\
\hline
Train/validation/test images & 102/22/22 & 117/15/14\\
Output channels & 2 & 11\\
Training crop & $512\times512$ & $512\times512$\\
Batch size & 1 & 1\\
Training epochs & 60 & 60\\
Mean epoch time (s) & 28.6 & 8.6\\
Best validation loss & 0.0384 & 0.0199\\
Validation Dice & 0.8896/0.8771 & 0.9578/0.9563/0.9733\\
\hline
\end{tabular}
\end{table}

The selected checkpoints were exported to FP32 ONNX and converted to TensorRT 10.3 FP16 and INT8 engines. Benchmarking used an NVIDIA RTX 3090. Six held-out images per task were used for the numerical comparison. Agreement Dice compares each TensorRT binary output with the PyTorch FP32 output on the same image. It is an implementation-equivalence measure and not a ground-truth segmentation score. Latency includes input normalization and inference but excludes the final GPU-to-CPU transfer.

\begin{table*}[t]
\caption{TensorRT benchmark}
\label{tab:seg_engine_benchmark}
\centering
\begin{tabular}{llcccc}
\hline
Task & Backend & Agreement Dice & Patch (ms) & Full frame (ms) & Size (MB)\\
\hline
Device & PyTorch FP32 & reference & 30.6 & 70.0 & --\\
Device & TensorRT FP16 & 0.9999/0.9997 & 8.1 & 32.8 & 60.1\\
Device & TensorRT INT8 & 0.9981/0.9854 & 4.2 & 17.1 & 32.1\\
Vessel & PyTorch FP32 & reference & 30.8 & 72.0 & --\\
Vessel & TensorRT FP16 & 0.9998 & 8.3 & 33.5 & 59.8\\
Vessel & TensorRT INT8 & 0.9812/0.9951 & 4.2 & 17.9 & 32.4\\
\hline
\end{tabular}
\end{table*}

Device agreement is listed as guidewire/catheter. Vessel FP16 reports the mean over eleven channels. Vessel INT8 reports mean/union agreement; its minimum branch-channel Dice is 0.9481. FP16 reduces single-patch latency by about $3.8\times$ and full-frame latency by about $2.1\times$. INT8 provides about $7.3\times$ and $4.1\times$ acceleration, respectively.

The mixed-precision deployment reflects how each segmentation error enters the navigation loop. Device masks are updated in every processed frame. Their thin guidewire and catheter structures directly determine tip position, local direction, route projection, and controller-state transitions. Device INT8 lowers full-frame latency from 32.8 to 17.1~ms, but catheter agreement decreases from 0.9997 to 0.9854. Device FP16 is therefore retained to reduce the propagation of quantization error into closed-loop decisions. Its measured 32.8-ms segmentation latency is below the nominal 66.7-ms acquisition interval associated with the requested 15-frame/s input. This benchmark covers normalization and segmentation inference only. It does not include geometric post-processing, state construction, communication, robot execution, Mentice updating, or image refresh.

Vessel inference is performed only during map initialization. It uses a ten-frame averaged input. The resulting vessel mask, centerline graph, branch labels, and planned route are then locked and confirmed by the operator. Consequently, vessel quantization error is not introduced into every subsequent feedback update. Vessel INT8 retains a union-channel agreement of 0.9951 while reducing full-frame latency from 33.5 to 17.9~ms and engine size from 59.8 to 32.4~MB. This operating point was selected for the static vessel representation. The choice does not imply that vascular geometry is less important than device tracking.

\subsection{Closed-loop experimental platform}
\label{sec:exp_platform}
\begin{figure*}[!t]
  \centering
  \includegraphics[width=0.90\textwidth]{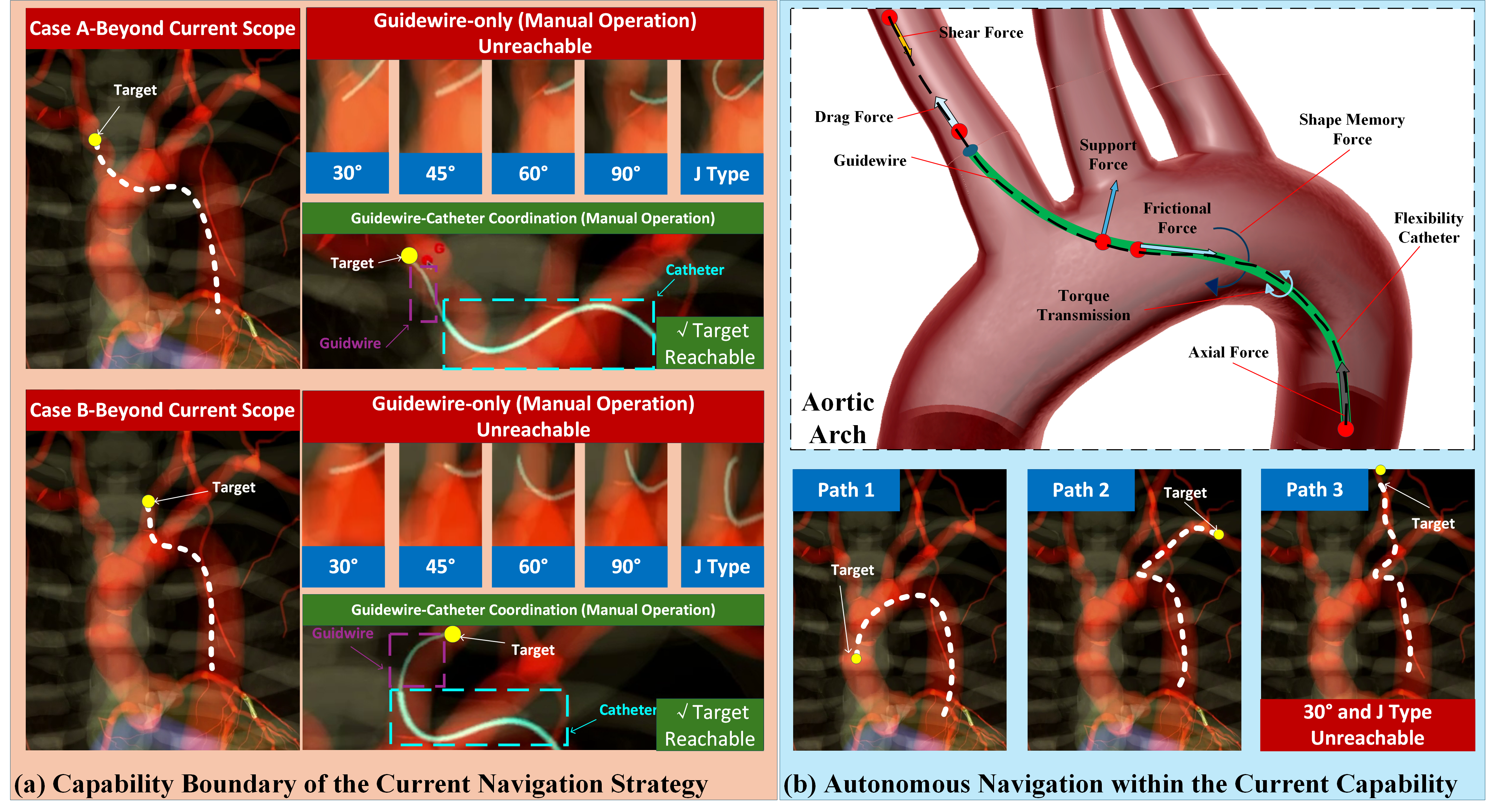}
  \caption{Capability boundary and evaluated navigation paths. (a) Manual guidewire-only tests define the conditions that can be completed before autonomous testing; the 30$^\circ$ and J guidewires do not complete Path 3 under the single-guidewire constraint, whereas the corresponding coordinated guidewire--catheter task remains reachable. (b) Simulator-rendered scenes for Paths 1--3 show the operator-selected start and target points and the planned vessel routes. Paths 1 and 2 include five guidewire conditions, while Path 3 includes only the three conditions that passed the prespecified capability check.}
  \label{fig:action}
\end{figure*}

\begin{figure*}[!t]
  \centering
  \includegraphics[width=0.90\textwidth]{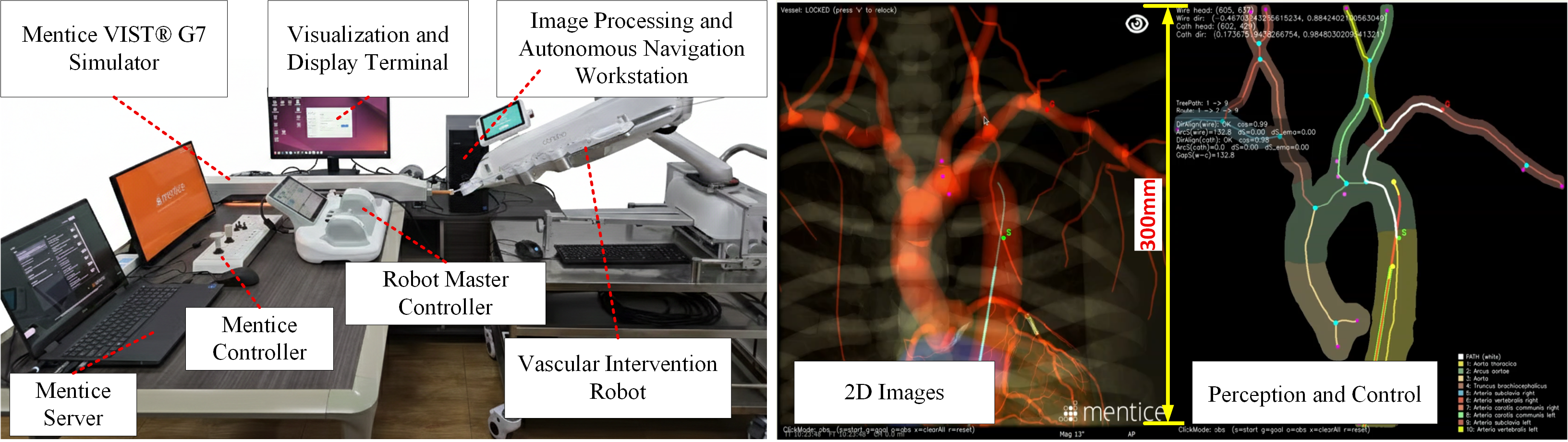}
  \caption{Closed-loop experimental platform. The platform combines the Mentice VIST G7 simulator, physical dual-device intervention robot, controller interfaces, display terminal, and autonomous-navigation workstation. It links device actuation, planar image generation, visual-state estimation, and autonomous commands in one supervised loop; the present dataset contains simulator-rendered projections rather than clinical DSA acquisitions.}
  \label{fig:platform}
\end{figure*}

To evaluate the image sensing and downstream path-guided control (Secs.~\ref{sec:visual_state}--\ref{sec:controller}), a closed-loop experimental platform was established that integrates (i) the Mentice vascular simulator and image source, (ii) a physical dual-device robotic actuation system, and (iii) a vision-based autonomous-navigation workstation. The platform permits the same simulator-rendered images to be assessed both as perception measurements and as inputs to the autonomous controller.
Fig.~\ref{fig:action} shows the retained capability-boundary and navigation scenes for the evaluated paths. The complete closed-loop workstation and actuation arrangement is shown in Fig.~\ref{fig:platform}.

\begin{figure*}[!t]
  \centering
  \includegraphics[width=0.96\textwidth]{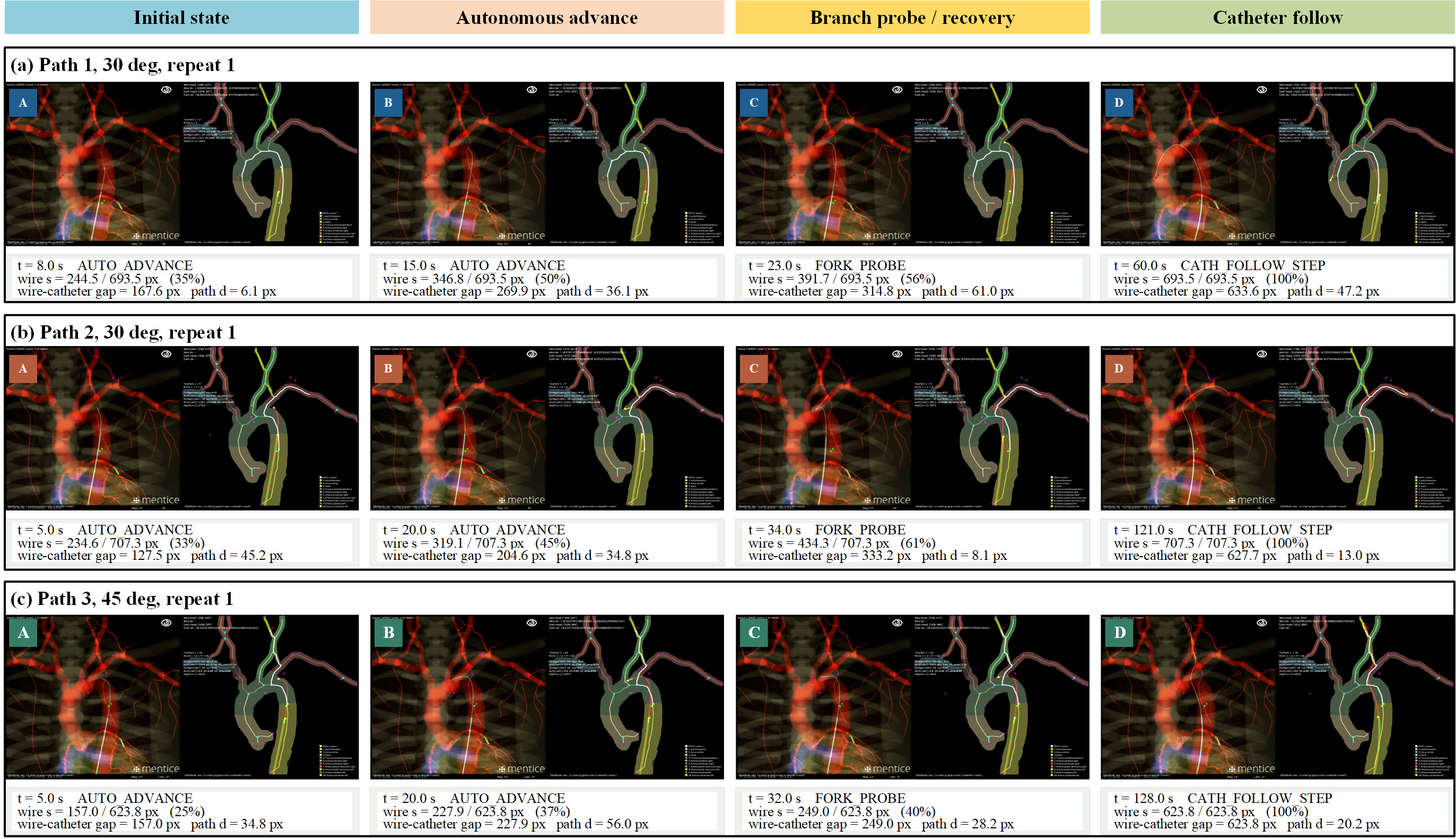}
  \caption{Representative autonomous navigation sequences across the three paths. Each row shows one recorded physical-robot run and follows the same four stages: initial state, autonomous advance, branch probing or recovery, and catheter-follow action. The values below each frame are taken from the synchronized action and state records.}
  \label{fig:three_path_process}
\end{figure*}

Figure~\ref{fig:action} documents the qualitative task boundary used before autonomous testing. The scene panels show the three locked routes and the marked start/target configuration; the manual capability panels show that the 30$^\circ$ and J guidewires could not complete Path 3 under the single-guidewire constraint. These two Path 3 conditions were excluded before the autonomous comparison, so their absence from the later tables and matrix is a design decision rather than a missing measurement. Figure~\ref{fig:platform} identifies the simulator, dual-device robot, rendered-image display, and navigation workstation used to close the loop. Figure~\ref{fig:three_path_process} then shows how the image-derived state drives representative autonomous actions along all three paths, including branch probing, recovery, and catheter-follow triggering. Together, the figures establish the geometry, hardware context, and recorded closed-loop process for the 65 logged trials.

\subsubsection{Task setup and protocol}
The operator selected the start, target, and interaction-risk points before the first trial on each path. The vessel topology, topology-constrained path, and marked points were then locked for five consecutive trials. Paths 1 and 2 were tested with 30$^\circ$, 45$^\circ$, 60$^\circ$, 90$^\circ$, and J guidewires (five repeats per condition; $n=25$ per path). During the manual capability check, the 30$^\circ$ and J guidewires could not complete Path 3 under the single-guidewire constraint; these two conditions were therefore excluded at the design level, and Path 3 was tested with 45$^\circ$, 60$^\circ$, and 90$^\circ$ guidewires (five repeats per condition; $n=15$). The exclusion was applied before the autonomous comparison and no values were imputed for the two unavailable Path 3 conditions.

At the start of each trial, the workstation updated $\mathbf{z}_t$ and executed Algorithm~\ref{alg:arc_length_nav}. The video and state export was complete through guidewire arrival and the catheter-follow trigger, but it was not guaranteed to contain the full downstream catheter endpoint sequence. Guidewire arrival was therefore summarized by the maximum observed route progress, with a prespecified reference of $\max_t(s^w/s_g)\geq0.95$. A catheter-follow trigger was counted when the logged controller entered the catheter-follow action after guidewire arrival. An autonomous stop event was identified from the logged transition corresponding to the video terminal message ``AUTO-ADV stall ... $\rightarrow$ STOP''; junction-probe and catheter-follow transitions were excluded. A run was classified as a failure when it contained three or more such autonomous stop events. The final signed route-coordinate gap $g^{wc}$ was retained only as a descriptive endpoint snapshot. The manual reset returned both devices toward the fixed start configuration between trials. The topology, path, and marked points remained unchanged. Reset actions were excluded from the trial records.

\subsubsection{Measured signals and evaluation criteria}
The logger records device positions, state relative to the planned path variables, controller modes, and executed actions. The primary sensing measures were the per-run median interframe mask IoU, tip-jitter RMS, direction cosine, and vision inference time for the guidewire and catheter. Interframe IoU quantifies temporal mask stability and is not a ground-truth segmentation accuracy. These measurements were linked to downstream process outcomes: maximum observed path-progress ratio $\max_t(s^w/s_g)$, the stop-based failure indicator (three or more autonomous stop events in one run), full run duration, action count, forward-action fraction, and the number of backward or turning actions. Catheter-follow triggering was reported as a binary process event. Action cadence was retained as a timing measure. The final-frame squared wire projection distance was converted to an image-space distance, and the final signed wire--catheter route-coordinate gap was retained for endpoint inspection only.

All statistics used the complete run as the independent unit. Frame-level measurements were reduced to one median per run; frames were not treated as independent replicates. Figures report individual runs together with the mean and descriptive percentile-bootstrap 95\% confidence interval. The full duration was computed from the run start to the recorded full-run endpoint, whereas action duration ended at the last logged action. This distinction is preserved in the reported values.

\subsection{Perception and closed-loop navigation results}
\label{sec:exp_quantitative_results}
Figs.~\ref{fig:action} and \ref{fig:three_path_process} provide the qualitative capability boundary and representative navigation sequences. The quantitative analysis below first examines the stability and timing of the image-derived state and then evaluates its downstream navigation behavior. It uses the complete action, frame, and group-summary logs from the same three paths. The route geometry is therefore interpreted through the same arc-length state used online, rather than through an independent post hoc path definition.

\begin{table*}[t]
\caption{Run-level outcomes.}
\label{tab:navigation_summary}
\centering
\scriptsize
\begin{tabular}{lrrrrrrrr}
\hline
Path & $n$ & Stops/run & Failed runs & Max. wire progress & Cath.-follow trigger & Full duration (s) & Actions & Forward fraction (\%) \\
\hline
Path 1 & 25 & $0.00\pm0.00$ & 0/25 & $1.000\pm0.000$ & 25/25 (100\%) & $111.7\pm37.7$ & $79.9\pm33.4$ & $76.0\pm11.6$ \\
Path 2 & 25 & $0.56\pm0.87$ & 2/25 & $1.000\pm0.000$ & 25/25 (100\%) & $162.6\pm75.4$ & $112.1\pm64.3$ & $71.2\pm10.6$ \\
Path 3 & 15 & $0.07\pm0.26$ & 0/15 & $1.000\pm0.000$ & 15/15 (100\%) & $148.2\pm34.3$ & $118.8\pm28.3$ & $66.2\pm5.0$ \\
\hline
\end{tabular}
\end{table*}

\begin{table}[H]
\caption{Imaging and timing.}
\label{tab:imaging_summary}
\centering
\scriptsize
\begin{tabular}{lccc}
\hline
Path & Wire IoU & Catheter IoU & Vision time (ms) \\
\hline
Path 1 & $0.890\pm0.013$ & $0.949\pm0.006$ & $70.5\pm0.9$ \\
Path 2 & $0.882\pm0.016$ & $0.943\pm0.012$ & $70.2\pm0.8$ \\
Path 3 & $0.897\pm0.013$ & $0.947\pm0.010$ & $71.1\pm0.7$ \\
\hline
\end{tabular}
\end{table}

\subsubsection{Route-level outcome and control effort}
\begin{figure}[H]
\centering
\includegraphics[width=0.82\columnwidth]{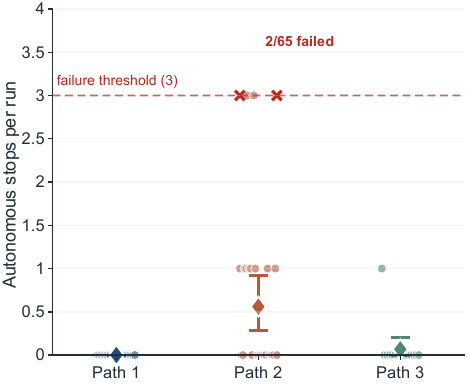}
\caption{Autonomous stop counts for individual runs. The dashed line marks the prespecified three-stop failure threshold; crosses identify the two failed runs.}
\label{fig:nav_stops}
\end{figure}

Figure~\ref{fig:nav_stops} shows the run-level stop distribution. All 65 runs reached at least 0.95 maximum observed guidewire progress, and all 65 logged the catheter-follow trigger. Two runs (3.1\%) were classified as failures: Path 2, 45$^\circ$, repeats 1 and 5, each contained three autonomous stop events. Paths 1 and 3 contained no failed runs; their largest stop count was one.

Path 2 had the largest mean duration and action count ($162.6\pm75.4$~s and $112.1\pm64.3$ actions), compared with $111.7\pm37.7$~s and $79.9\pm33.4$ actions for Path 1. Path 3 required $148.2\pm34.3$~s and $118.8\pm28.3$ actions. The matrix in Fig.~\ref{fig:nav_matrix} shows the long Path 2 repeats directly: 431~s/341 actions for the 30$^\circ$ fifth repeat and 295~s/239 actions for the 90$^\circ$ second repeat.

\subsubsection{Strategy composition and guidewire dependence}
\begin{figure}[H]
\centering
\includegraphics[width=0.92\columnwidth]{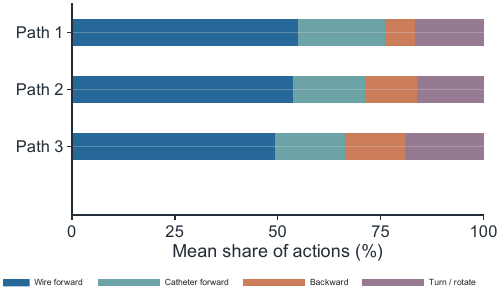}
\caption{Mean per-run composition of guidewire advance, catheter advance, backward movement, and turning or rotation across the three paths.}
\label{fig:nav_actions}
\end{figure}

Figure~\ref{fig:nav_actions} shows the mean action composition. Forward actions accounted for 76.0\%, 71.2\%, and 66.2\% on Paths 1--3. Backward actions accounted for 7.3\%, 12.7\%, and 14.6\%, while turns accounted for 16.7\%, 16.2\%, and 19.2\%. The lower forward share on Path 3 coincided with more backward and turning actions. Fractions are averaged per run.

\begin{figure*}[!t]
\centering
\includegraphics[width=0.72\textwidth]{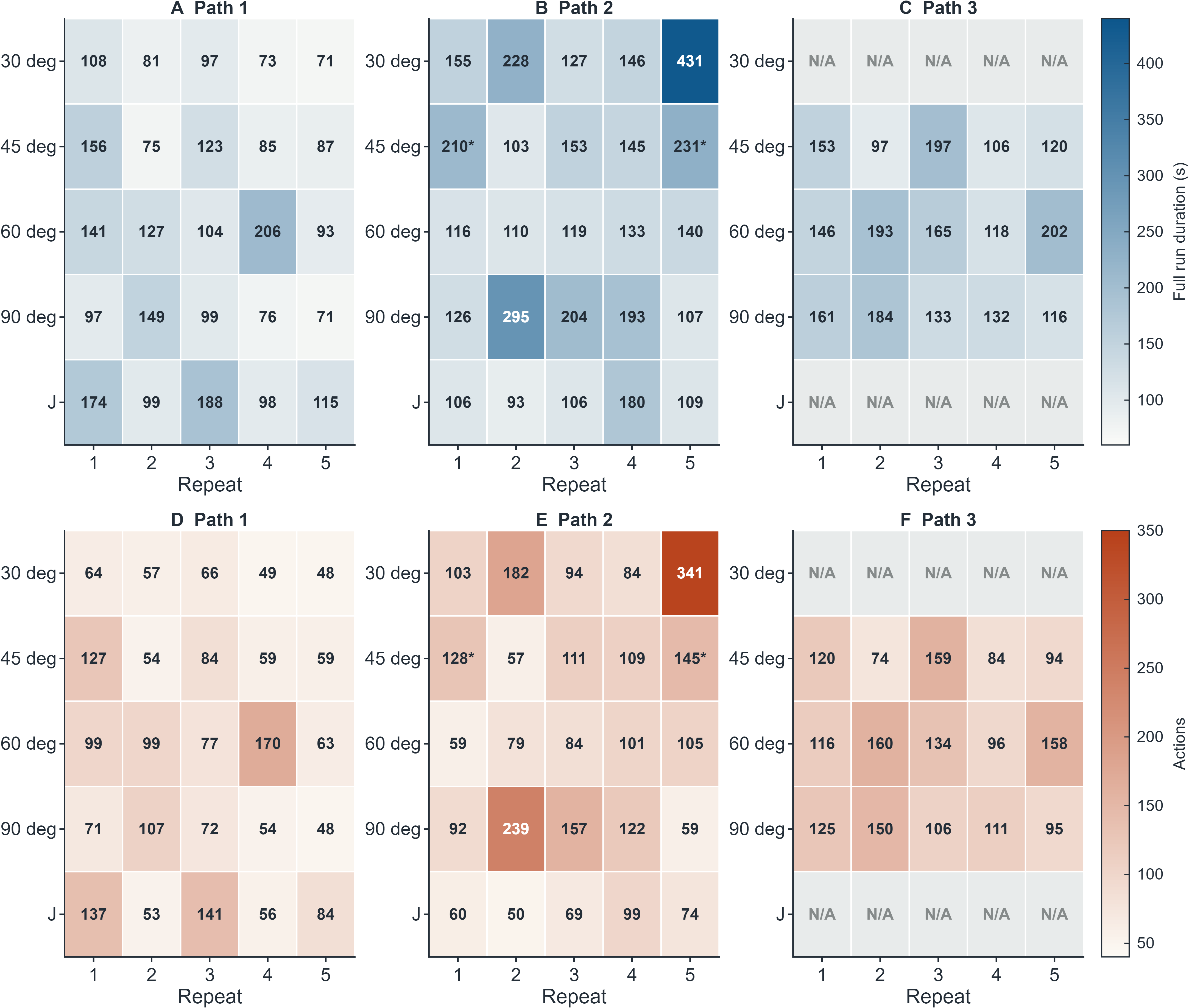}
\caption{Complete trial matrices. Panels A--C show full run duration and panels D--F show action count; each cell is one logged repeat. Asterisks mark the two stop-based failures and N/A denotes the excluded Path 3 conditions.}
\label{fig:nav_matrix}
\end{figure*}

Figure~\ref{fig:nav_matrix} gives the repeat-level records. Panels A--C show duration and panels D--F action count for the same path/guidewire/repeat grid. Asterisks identify the two failed Path 2 45$^\circ$ repeats. The Path 3 30$^\circ$ and J cells are N/A because those conditions failed the prespecified manual capability check.

Guidewire-condition values are summarized in Table~\ref{tab:condition_summary}. On Path 1, the 30$^\circ$ condition was the shortest on average (86.1~s) and the 60$^\circ$ condition required the most actions (101.6). Path 2 was most demanding for the 30$^\circ$ guidewire (217.5~s and 160.8 actions), whereas the J condition was shorter (118.8~s and 70.4 actions). On Path 3, the 60$^\circ$ guidewire required the largest mean effort among the available conditions (164.8~s and 132.8 actions). The 30$^\circ$ and J cells for Path 3 are intentionally absent because those conditions were excluded by the prespecified manual capability boundary; no cross-path comparison is made for those cells.

\begin{table*}[!b]
\caption{Condition-level summary.}
\label{tab:condition_summary}
\centering
\scriptsize
\begin{tabular}{llrrrrr}
\hline
Path & Guidewire & $n$ & Duration (s) & Actions & Forward (\%) & Total stops; failed runs \\
\hline
1 & 30$^\circ$ & 5 & $86.1\pm16.0$ & $56.8\pm8.3$ & $87.3\pm3.2$ & 0; 0 \\
1 & 45$^\circ$ & 5 & $105.3\pm33.7$ & $76.6\pm30.5$ & $75.6\pm8.6$ & 0; 0 \\
1 & 60$^\circ$ & 5 & $134.0\pm44.3$ & $101.6\pm41.2$ & $66.7\pm11.4$ & 0; 0 \\
1 & 90$^\circ$ & 5 & $98.5\pm31.0$ & $70.4\pm23.0$ & $77.5\pm9.4$ & 0; 0 \\
1 & J & 5 & $134.6\pm43.0$ & $94.2\pm42.7$ & $73.1\pm14.9$ & 0; 0 \\
2 & 30$^\circ$ & 5 & $217.5\pm125.6$ & $160.8\pm108.0$ & $66.1\pm8.3$ & 3; 0 \\
2 & 45$^\circ$ & 5 & $168.3\pm51.4$ & $110.0\pm33.0$ & $68.9\pm10.6$ & 8; 2 \\
2 & 60$^\circ$ & 5 & $123.4\pm12.3$ & $85.6\pm18.5$ & $72.9\pm8.4$ & 1; 0 \\
2 & 90$^\circ$ & 5 & $185.1\pm74.3$ & $133.8\pm69.1$ & $66.2\pm10.0$ & 1; 0 \\
2 & J & 5 & $118.8\pm34.7$ & $70.4\pm18.4$ & $81.7\pm10.3$ & 1; 0 \\
3 & 45$^\circ$ & 5 & $134.8\pm40.6$ & $106.2\pm34.1$ & $68.7\pm6.7$ & 1; 0 \\
3 & 60$^\circ$ & 5 & $164.8\pm34.3$ & $132.8\pm27.4$ & $63.7\pm3.9$ & 0; 0 \\
3 & 90$^\circ$ & 5 & $145.1\pm26.8$ & $117.4\pm21.2$ & $66.3\pm3.5$ & 0; 0 \\
\hline
\end{tabular}
\end{table*}

\subsubsection{Imaging stability and path-relative state quality}
Table~\ref{tab:imaging_summary} reports the sensing measures. Median wire-mask IoU was 0.890, 0.882, and 0.897 for Paths 1--3; catheter-mask IoU was 0.949, 0.943, and 0.947. Vision time varied little (70.2--71.1~ms). Wire-tip jitter was approximately 3 pixels and catheter-tip jitter 2.5--2.7 pixels. Catheter direction cosine stayed near 0.99, whereas wire direction cosine was lower and more variable (0.583, 0.347, and 0.335), as expected during retreat and branch reorientation.

\section{Discussion}
\label{sec:discussion}
The path-relative state remained available throughout the tested robot runs. Wire-mask IoU ranged from 0.882 to 0.897, catheter-mask IoU from 0.943 to 0.949, and vision time from 70.2 to 71.1~ms. Tip jitter was approximately 3 pixels for the wire and 2.5--2.7 pixels for the catheter. The lower and more variable wire direction cosine reflects the reorientation and retreat states encountered during navigation. These are deployment measurements; they do not establish localization accuracy on clinical fluoroscopy.

The state also exposed the reason for each control action. Retraction, rotation scans, insertion tests, rollback, risk retreat, and catheter following were tied to explicit thresholds or mode transitions. All 65 runs reached at least 0.95 maximum observed guidewire progress, but two Path 2 45$^\circ$ runs reached the three-stop failure threshold. The matrices show that the same state variables remained available during the longest runs.

Mechanical contact and friction still redirect the flexible guidewire during rotation and translation. The junction routine represents this interaction as a bounded trial, and matched rollback reverses the executed insertion count after an unacceptable outcome. Risk-point recovery provides an operator-marked early retreat, whereas off-path recovery reacts to persistent projection deviation. Neither mechanism measures contact force or constitutes a formal safety layer.

The study used three routes, 65 physical-robot runs, one fixed projection, and one parameter set. Path 3 excludes the 30$^\circ$ and J conditions after the single-guidewire capability check, so those cells cannot support balanced cross-path comparisons. The segmentation models were trained on modest simulated-image datasets, and their Dice and TensorRT measurements should not be transferred to clinical fluoroscopy without further testing. Pixel thresholds, view-dependent turning, a locked vessel map, and tip-availability-based detection remain experimental assumptions. Recording ended after guidewire arrival and catheter-follow triggering; the final wire--catheter gap is therefore an endpoint snapshot, not a catheter-completion label. Calibrated localization, complete two-device endpoint trials, broader anatomy, clinical image sequences, and anatomically realistic benchtop or preclinical models remain to be evaluated.

Future work will extend the state to local guidewire--catheter coordination and to image sequences with changing projection and anatomy.

\section{Conclusion}
\label{sec:conclusion}
Autonomous navigation under planar image feedback can be formulated as a state-construction problem. The presented formulation separates static vessel topology from dynamic device motion and expresses the latter in a route-relative coordinate system: the topology graph retains branch identity, arc length provides continuous progress, and geometric residuals expose deviation, direction consistency, and guidewire--catheter lag. Together, these variables form an interpretable state for route-level decisions under incomplete visual observations.

On the physical dual-device robot, the state supported target-directed delivery, branch probing, rollback, recovery, and catheter-follow triggering across 65 closed-loop runs. All runs reached at least 0.95 maximum observed guidewire progress, and two Path 2 runs met the prespecified three-stop failure criterion. The evidence supports a path-relative image state as a practical abstraction for traceable macroscopic navigation under a fixed planar projection. Clinical image sequences, changing projections, calibrated localization, force-related state variables, broader anatomy, and complete catheter endpoint trials remain necessary to establish generalization and clinical relevance.

\begin{IEEEbiography}[{\includegraphics[width=0.68in]{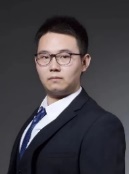}}]{Sheng Cao (Student Member, IEEE)}
received the B.S. degree in Power System and Automation from Army Engineering University of PLA, Nanjing, Jiangsu Province, China, in 2018 and the M.S. degree in Control Science and Engineering from Tianjin University of Technology, Tianjin, China, in 2021. He is currently pursuing the Ph.D. degree with Beijing Institute of Technology, Beijing, China.
\end{IEEEbiography}

\begin{IEEEbiography}[{\includegraphics[width=0.68in]{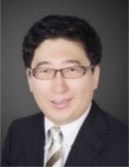}}]{Shuxiang Guo (Fellow, IEEE)}
received the Ph.D. degree in Mechano-Informatics and Systems from Nagoya University, Japan, in 1995. He is currently a Chair Professor with the Key Laboratory of Convergence Medical Engineering System and Healthcare Technology, Ministry of Industry and Information Technology, Beijing Institute of Technology, Beijing, China.
\end{IEEEbiography}

\begin{IEEEbiography}[{\includegraphics[width=0.68in]{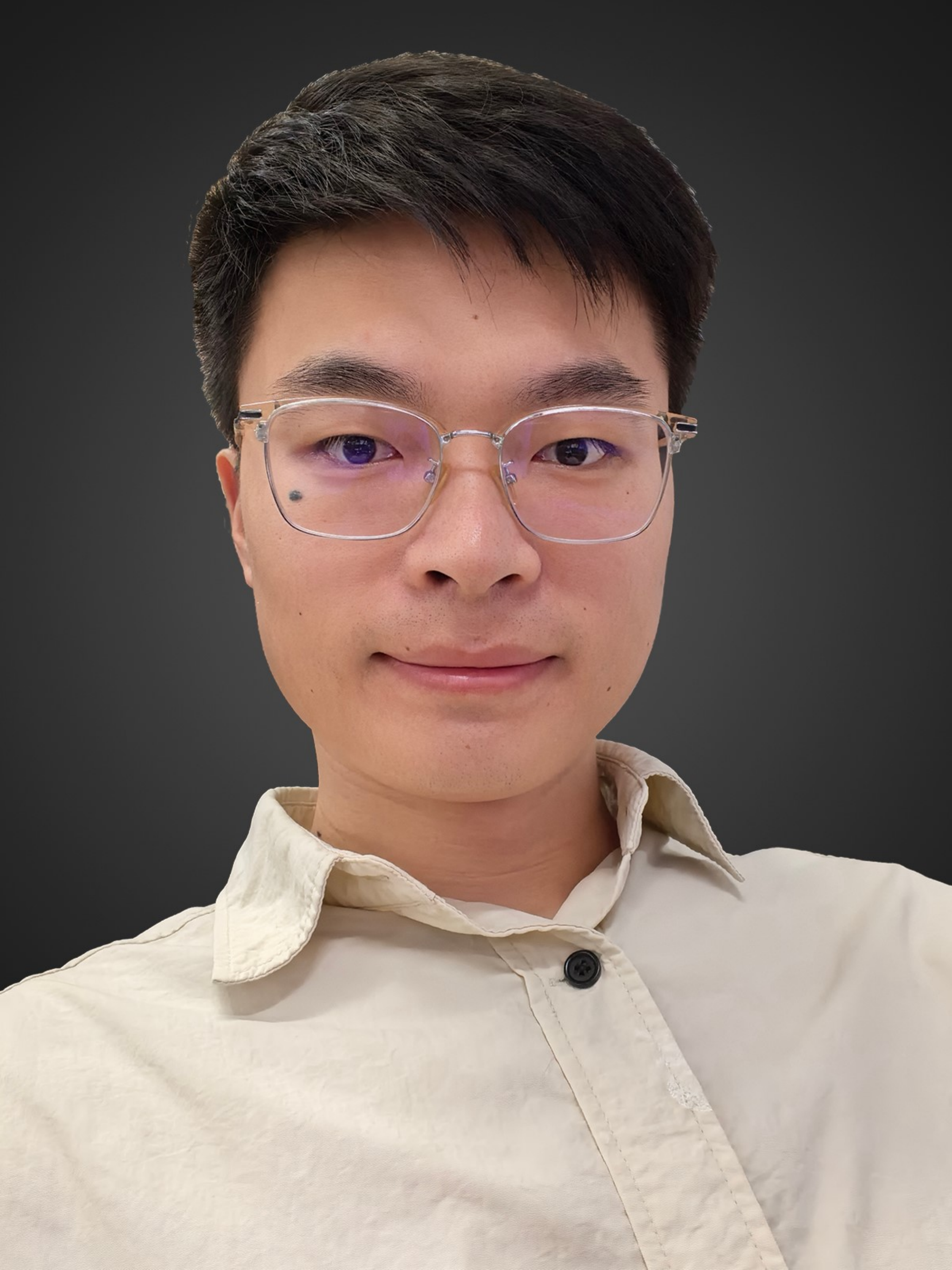}}]{Yonggan Yan (Member, IEEE)}
received the Ph.D. degree in biomedical engineering from Beijing Institute of Technology, Beijing, China, in 2025. He is currently a Lecturer with the School of Mechanical Electronic and Control Engineering, Beijing Jiaotong University, Beijing, China. His research interests include surgical robots and magnetic navigation.
\end{IEEEbiography}

\begin{IEEEbiography}[{\includegraphics[width=0.68in]{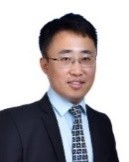}}]{Jian Guo}
received the Ph.D. degree in Intelligent Mechanical Systems Engineering from Kagawa University, Japan, in 2012. He is currently a part-time Professor with the School of Electrical Engineering and Automation, Tianjin University of Technology, Tianjin, China, and the CEO of Shenzhen Institute of Advanced Biomedical Robot Co., Ltd., Shenzhen, China.
\end{IEEEbiography}

\begin{IEEEbiography}[{\includegraphics[width=0.68in]{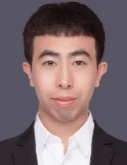}}]{Chunying Li}
received the Ph.D. degree from the Graduate School of Engineering, Kagawa University, Takamatsu, Japan, in 2023. He is currently a Research Assistant Professor with the Advanced Institute for Ocean Research, Southern University of Science and Technology, Shenzhen, China. His research interests include bionic underwater robots, multisensor systems, and multirobot collaboration.
\end{IEEEbiography}

\newpage
\begin{IEEEbiography}[{\includegraphics[width=0.68in]{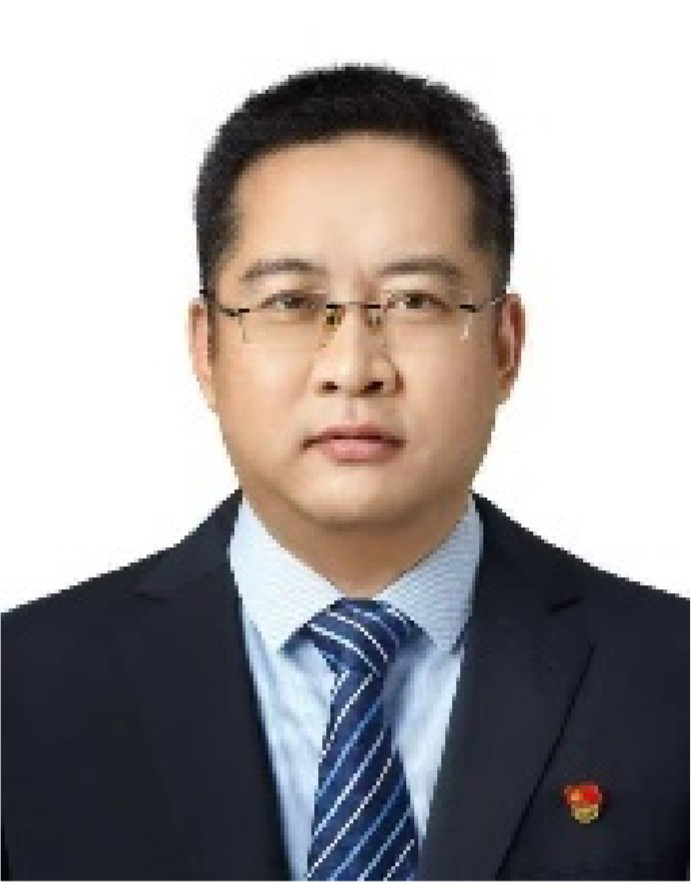}}]{Bin Wang}
is a Director of the Interventional Vascular Department at Aerospace Center Hospital, Beijing, China, a leader in the subspecialty of peripheral vascular disease intervention, a Chief Physician, and a Master's Supervisor. He is proficient in the interventional treatment of lower-limb deep-vein thrombosis and pulmonary embolism, acute and chronic ischemic peripheral vascular diseases, and hemorrhagic diseases requiring interventional embolization.
\end{IEEEbiography}

\begin{IEEEbiography}[{\includegraphics[width=0.68in]{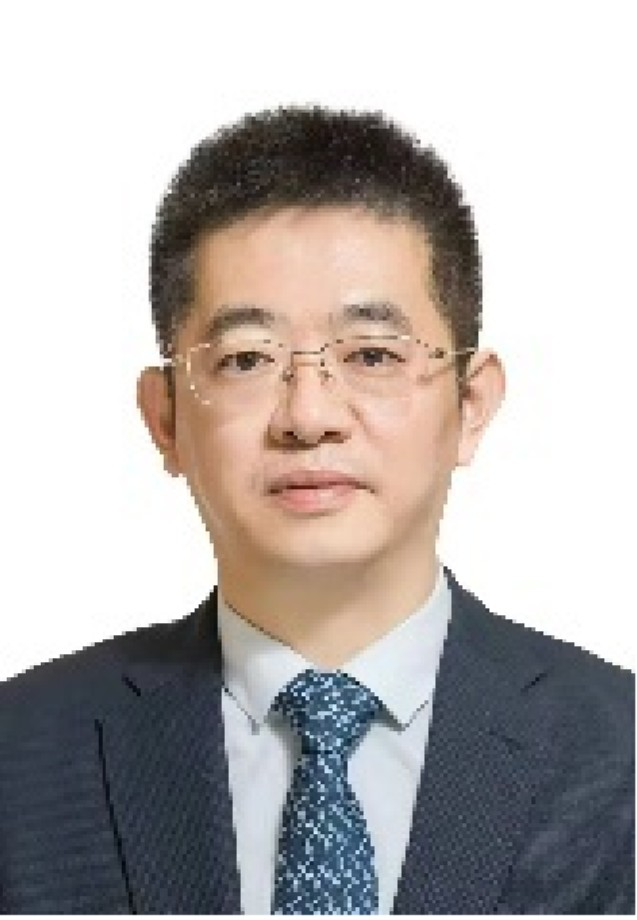}}]{Mingchao Ding}
received the M.S. degree in imaging and nuclear medicine from Peking University School of Medicine, Beijing, China, in 2011. He is currently the President, Chief Physician, Professor, and Doctoral Supervisor of Aerospace Center Hospital, and leads the Continuing Education and Training Base for Peripheral Vascular Interventional Physicians of the National Health Commission. His research focuses on endovascular and emergency interventional treatment.
\end{IEEEbiography}

\begin{IEEEbiography}[{\includegraphics[width=0.68in]{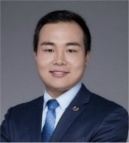}}]{Yongxin Zhang}
received the Doctor of Philosophy and medical doctor degrees from the Second Military Medical University, China, in 2017. He is currently the Secretary of the OCIN Imaging Core Laboratory. His representative papers have been published in \emph{The New England Journal of Medicine} and \emph{The Lancet}.
\end{IEEEbiography}

\begin{IEEEbiography}[{\includegraphics[width=0.68in]{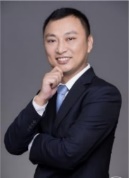}}]{Yongwei Zhang}
received the master's degree from the Second Military Medical University, China, in 2011. He has twenty years of experience in neurointerventional treatment and was one of the early surgeons in the interventional treatment of acute ischemic stroke in China. He is currently the Vice Director of the Neurovascular Center, Shanghai Changhai Hospital, Shanghai, China.
\end{IEEEbiography}

\begin{IEEEbiography}[{\includegraphics[width=0.68in]{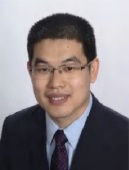}}]{Pengfei Yang}
completed the seven-year program in clinical medicine at Zhejiang University School of Medicine, China, in 2007 and received the M.D. degree from the Second Military Medical University, China, in 2013. He is currently the Executive Director of the Neurovascular Center, Shanghai Changhai Hospital, and the Vice Director of the Institute of Cerebrovascular Diseases, PLA.
\end{IEEEbiography}

\begin{IEEEbiography}[{\includegraphics[width=0.68in]{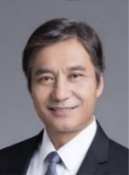}}]{Jianmin Liu}
received the bachelor's degree from the Second Military Medical University, China, in 1984 and was a visiting professor at Osaka University Medical College, Japan, in 1997. He is currently the Director of the Neurovascular Center, Shanghai Changhai Hospital, and the Director of the Institute of Cerebrovascular Disease, PLA.
\end{IEEEbiography}

\end{document}

%% file: intro.tex
\IEEEPARstart{E}{ndovascular} intervention is a sequential decision problem observed through a planar projection. The clinical need for reliable vascular intervention has motivated robotic assistance across vascular and neuroendovascular procedures \cite{heart_statistics_2026,vascular_robot_review_2024,cardiovascular_robot_review_2025,neuro_robotics_2025}. Recent studies have also examined autonomous endovascular systems and embodied intelligence for image-guided intervention \cite{autonomous_endovascular_review_2025,embodied_endovascular_review_2026}. During DSA or X-ray fluoroscopy, vessel anatomy, device--vessel contact, and depth are represented only indirectly in the image. Each device command changes the configuration that produces the next image, so autonomous navigation requires a state that remains useful under incomplete and changing observations.

Local image cues do not uniquely determine route-level intent. Existing fluoroscopic methods have addressed catheter segmentation and guidewire-pose tracking, but these measurements do not by themselves specify the intended route or the next action \cite{catheter_seg_fluoro_2017,guidewire_pose_fluoro_2023}. A guidewire tip and its local tangent can be consistent with more than one branch, while projection overlap, occlusion, and segmentation jitter can change the apparent direction. A useful state must therefore preserve branch identity and progress along the selected route while suppressing pixel-level variation that is irrelevant to the next decision.

The path-relative representation used here provides this task-oriented abstraction. A vessel centerline is represented as a labeled topology graph, and the operator-selected route is parameterized by arc length. Projection of the guidewire and catheter tips onto the route yields progress, transverse deviation, directional consistency, and inter-device lag. The graph supplies discrete branch structure; arc length supplies continuous progress; and the residuals indicate whether the observation remains compatible with the intended motion. Static vessel topology is initialized and locked, whereas device measurements remain frame dependent. This factorization converts image observations into a state that can be inspected at every control transition.

Image-guided navigation studies have used local directional cues, learned policies, and short-horizon visual feedback \cite{rl_zeroshot_guidewire_2024,rl_pathplanning_guidewire_2024,safe_dual_device_rl_2025}. Route progress, branch identity, and recovery conditions are not always explicit in these control states. A deterministic policy based on path-relative variables makes the cause of each advancement, probe, rollback, or recovery transition observable and allows run-level failures to be traced to image-derived conditions.

The formulation targets the macroscopic layer of endovascular autonomy: delivery along a selected route, simple branch choice, deviation correction, and recovery. The physical dual-device robot builds on prior multi-manipulator, autonomous-delivery, remote-control, and multi-device intervention platforms \cite{cao_reciprocating_robot_2024,cao_autonomous_delivery_2025,cao_webrtc_2025,zhang_multidevice_2026}. Force-related sensing, multimodal feedback, haptic interfaces, and steerable-catheter mechanisms provide complementary directions for extending the control state \cite{lyu_force_sensing_2024,li_multimodal_feedback_2025,mei_haptic_2026,yan_steerable_catheter_2025}. The Mentice platform provides repeatable planar projections for controlled validation. Two segmentation models, skeleton-based tip and direction estimation, topology-constrained route construction, and a deterministic state machine form the closed loop. At a junction, an alternating rotation scan is followed by a bounded advancement trial; the next image determines acceptance or rollback. Direct force measurement and three-dimensional reconstruction are not required for this route-level formulation.

The central contribution is a topology-constrained, path-relative state abstraction that converts planar device and vessel observations into interpretable progress and branch-decision variables for physical-robot navigation. Route-level behavior is evaluated over 65 runs on three paths, with explicit stop-based failure accounting and image-derived measurements reported throughout the control loop.

%% file: root.bbl
\begin{thebibliography}{99}

\bibitem{heart_statistics_2026}
L.~P.~Palaniappan, \emph{et al.},
``2026 heart disease and stroke statistics: A report of US and global data from the American Heart Association,''
\emph{Circulation}, vol.~153, no.~9, pp.~e275--e906, Mar.~2026,
doi: 10.1161/CIR.0000000000001412.

\bibitem{vascular_robot_review_2024}
B.~C.~Lengyel, P.~Chinnadurai, S.~J.~Corr, et al.,
``Robot-assisted vascular surgery: Literature review, clinical applications, and future perspectives,''
\emph{J.\ Robot.\ Surg.}, vol.~18, no.~1, p.~328, 2024.


\bibitem{cardiovascular_robot_review_2025}
S.~R.~Mandhumula and M.~A.~Tavallaei,
``Robotic systems in cardiovascular interventions: Evolving platforms and the path forward,''
\emph{Methodist DeBakey Cardiovasc.~J.}, vol.~21, no.~5, pp.~16--34, 2025,
doi: 10.14797/mdcvj.1652.

\bibitem{neuro_robotics_2025}
S.~Miyachi, Y.~Nagano, R.~Kawaguchi, and H.~Tadauchi,
``Robotics for neuroendovascular therapy,''
\emph{J.\ Neuroendovascular Ther.}, vol.~19, no.~1, pp.~ra.2024-0112, 2025,
doi: 10.5797/jnet.ra.2024-0112.

\bibitem{autonomous_endovascular_review_2025}
J.~Sun, R.~Tang, Q.~Zhang, \emph{et al.},
``Advancements and challenges in autonomous endovascular interventional robotics: A comprehensive review,''
\emph{iScience}, vol.~28, no.~12, Art.~no.~114024, 2025,
doi: 10.1016/j.isci.2025.114024.

\bibitem{embodied_endovascular_review_2026}
T.~Yao, \emph{et al.},
``Advancing embodied intelligence in robotic-assisted endovascular procedures: A systematic review of AI solutions,''
\emph{IEEE Rev.~Biomed.~Eng.}, vol.~19, pp.~248--266, 2026,
doi: 10.1109/RBME.2025.3641383.

\bibitem{catheter_seg_fluoro_2017}
P.~Ambrosini, D.~Ruijters, W.~J.~Niessen, A.~Moelker, and T.~van~Walsum,
``Fully automatic and real-time catheter segmentation in X-ray fluoroscopy,''
in \emph{Proc.\ MICCAI}, 2017, pp.~577--585.


\bibitem{guidewire_pose_fluoro_2023}
S.~R.~Ravigopal, A.~Sarma, T.~A.~Brumfiel, and J.~P.~Desai,
``Real-time pose tracking for a continuum guidewire robot under fluoroscopic imaging,''
\emph{IEEE Trans.\ Med.\ Robot.\ Bionics}, vol.~5, no.~2, pp.~230--241, May 2023,
doi: 10.1109/TMRB.2023.3260273.

\bibitem{rl_zeroshot_guidewire_2024}
V.~Scarponi, M.~Duprez, F.~Nageotte, et al.,
``A zero-shot reinforcement learning strategy for autonomous guidewire navigation,''
\emph{Int.\ J.\ Comput.\ Assist.\ Radiol.\ Surg.}, vol.~19, no.~6, pp.~1185--1192, 2024.


\bibitem{rl_pathplanning_guidewire_2024}
W.~Liu, W.~Xu, X.~Li, et al.,
``Autonomous guidewire navigation for robot-assisted endovascular interventions: A knowledge-driven visual guidance approach,''
\emph{arXiv preprint arXiv:2403.05748}, 2024.


\bibitem{safe_dual_device_rl_2025}
H.~Robertshaw, B.~Jackson, J.~Wang, et al.,
``Reinforcement learning for safe autonomous two-device navigation of cerebral vessels in mechanical thrombectomy,''
\emph{Int.\ J.\ Comput.\ Assist.\ Radiol.\ Surg.}, vol.~20, no.~6, pp.~1077--1086, 2025.


\bibitem{cao_reciprocating_robot_2024}
S.~Cao, S.~Guo, J.~Guo, J.~Wang, \emph{et al.},
``A reciprocating delivery device-based endovascular intervention robot with multimanipulators collaboration,''
\emph{IEEE Trans.\ Instrum.\ Meas.}, vol.~73, pp.~1--12, 2024,
doi: 10.1109/TIM.2023.3342853.

\bibitem{cao_autonomous_delivery_2025}
S.~Cao, S.~Guo, J.~Guo, J.~Wang, C.~Li, H.~Xu, B.~Wang, and M.~Ding,
``Design of an autonomous delivery system for vascular intervention robots based on curvature variation,''
in \emph{Proc.~IEEE Int.~Conf.~Mechatronics Autom. (ICMA)}, 2025, pp.~104--109.

\bibitem{cao_webrtc_2025}
S.~Cao, S.~Guo, J.~Guo, J.~Wang, and J.~Ge,
``WebRTC and 5G based remote control system for a vascular intervention robot,''
in \emph{Proc.~IEEE/RSJ Int.~Conf.~Intell.~Robots Syst. (IROS)}, 2025, pp.~20624--20629,
doi: 10.1109/IROS60139.2025.11247408.

\bibitem{zhang_multidevice_2026}
Y.~Zhang, H.~Ma, J.~Li, \emph{et al.},
``Feasibility of a novel multi-device robotic system for multiple neurointerventional procedures: Pre-clinical validation in porcine models,''
\emph{J.~Robot.~Surg.}, vol.~20, no.~1, 2026, PMID: 42627538.

\bibitem{lyu_force_sensing_2024}
C.~Lyu, \emph{et al.},
``Deep learning-based force sensing method for a flexible endovascular surgery robot,''
\emph{IEEE Trans.\ Instrum.\ Meas.}, vol.~73, pp.~1--10, 2024,
doi: 10.1109/TIM.2024.3381290.

\bibitem{li_multimodal_feedback_2025}
X.~Li, S.~Guo, M.~Kawanishi, and K.~Suzuki,
``A multimodal feedback sensing system for an endovascular surgery robot,''
\emph{IEEE Sensors J.}, vol.~25, no.~16, pp.~31461--31472, 2025,
doi: 10.1109/JSEN.2025.3587324.

\bibitem{mei_haptic_2026}
Z.~Mei, S.~Han, S.~Qu, \emph{et al.},
``Design of an active haptic interface using proprioception feedback for continuous endovascular teleoperation,''
\emph{IEEE Robot.~Autom.~Lett.}, vol.~11, no.~6, pp.~7508--7515, Jun.~2026,
doi: 10.1109/LRA.2026.3688093.

\bibitem{yan_steerable_catheter_2025}
Y.~Yan, S.~Guo, H.~Shen, B.~Wang, and M.~Ding,
``A novel dexterous steerable catheter robot system: Design, modeling and evaluation,''
\emph{IEEE Robot.\ Autom.\ Lett.}, vol.~10, no.~9, pp.~8770--8777, Sept.~2025,
doi: 10.1109/LRA.2025.3588417.

\bibitem{saderi_registration_2025}
M.~Saderi, J.~H.~Patel, C.~D.~Sheagren, \emph{et al.},
``3D CT to 2D X-ray image registration for improved visualization of tibial vessels in endovascular procedures,''
\emph{Int.~J.~Comput.~Assist.~Radiol.~Surg.}, vol.~20, no.~4, pp.~753--763, 2025,
doi: 10.1007/s11548-024-03302-z.

\bibitem{ye_registration_2025}
W.~Ye, J.~Wu, W.~Zhang, \emph{et al.},
``A robust method for real-time intraoperative 2D and preoperative 3D X-ray image registration based on an enhanced Swin Transformer framework,''
\emph{Bioengineering}, vol.~12, no.~2, Art.~no.~114, 2025,
doi: 10.3390/bioengineering12020114.

\end{thebibliography}
